\documentclass{llncs}
\usepackage{graphicx} % for pdf, bitmapped graphics files
\usepackage{makeidx}  % allows for indexgeneration
\usepackage{amsmath}
\usepackage{amssymb}
\usepackage{subfigure}
\usepackage{mathtools}
\usepackage{xparse}
\usepackage{cite}
\usepackage{booktabs}
\usepackage{adjustbox}
\usepackage{pdflscape}
\usepackage{bm}
\usepackage{afterpage}
\usepackage{color}
\usepackage[title]{appendix}
\usepackage{setspace}

\usepackage[belowskip=-10pt,aboveskip=0pt]{caption}
\newcommand{\ra}[1]{\renewcommand{\arraystretch}{#1}}

\DeclarePairedDelimiter{\norm}{\lVert}{\rVert}
\NewDocumentCommand{\normL}{ s O{} m }{%
	\IfBooleanTF{#1}{\norm*{#3}}{\norm[#2]{#3}}_{L_2(\Omega)}%
}

\def\R{{\mathbb R}}        % reals
\newtheorem{assumption}{Assumption}

\begin{document}
\linespread{0.85}

\frontmatter          % for the preliminaries
\pagestyle{headings}  % switches on printing of running heads
\mainmatter              % start of the contributions
%%%%%%%%%%%%%%%%%%%%%%%%%%%%%%%%
% 									TITLE
%%%%%%%%%%%%%%%%%%%%%%%%%%%%%%%%
\title{Density Matching Reward Learning}
\titlerunning{DMRL}  % abbreviated title (for running head)  also used for the TOC unless \toctitle is used
\author{Sungjoon Choi\inst{1} \and Kyungjae Lee\inst{1}
\and H. Andy Park \inst{2} \and Songhwai Oh\inst{1}}
\authorrunning{Choi et al.} % abbreviated author list (for running head)
\institute{Seoul National University, Seoul, Korea\\
\email{\{sungjoon.choi, kyungjae.lee, songhwai.oh\}@cpslab.snu.ac.kr} 
\and
Rethink Robotics, Boston, MA, USA\\
\email{apark@rethinkrobotics.com}
}

\maketitle              % typeset the title of the contribution

%%%%%%%%%%%%%%%%%%%%%%%%%%%%%%%%
% 									ABSTRACT
%%%%%%%%%%%%%%%%%%%%%%%%%%%%%%%%
\begin{abstract} 
In this paper, we focus on the problem of inferring the 
underlying reward function of an expert given demonstrations,
which is often referred to as inverse reinforcement learning (IRL).
In particular, we propose a model-free density-based IRL algorithm,
named density matching reward learning (DMRL),
which does not require model dynamics.
% as it does not require to solve a Markov decision process as a sub-routine.
%% This \textit{model-free} property markedly increases the applicability
%% of the IRL problems.
The performance of DMRL is analyzed theoretically and the
sample complexity is derived.
Furthermore, the proposed DMRL is extended to handle nonlinear IRL
problems by assuming that the reward function is in the reproducing kernel
Hilbert space (RKHS) and kernel DMRL (KDMRL) is proposed. 
The parameters for KDMRL can be computed analytically, which greatly
reduces the computation time. 
The performance of KDMRL is extensively evaluated 
in two sets of experiments: 
grid world and track driving experiments.  
In grid world experiments, the proposed KDMRL method is compared with
both model-based 
%% MMP \cite{ratliff2006maximum},
%% MaxEnt \cite{ziebart2008maximum}, and
%% GPIRL \cite{levine2011nonlinear},
and model-free
%% RelEnt \cite{boularias2011relative},
IRL methods and shows superior performance on a nonlinear reward
setting and competitive performance on a linear reward setting
in terms of expected value differences. 
Then we move on to more realistic experiments of learning different
driving styles for autonomous navigation in complex and dynamic tracks
using KDMRL and receding horizon control. 
\end{abstract}

%%%%%%%%%%%%%%%%%%%%%%%%%%%%%%%%
% 									INTRODUCTION
%%%%%%%%%%%%%%%%%%%%%%%%%%%%%%%%
\section{Introduction}

% 1. Emphasis on skill learning / why? / Applications / RL? 
Modern autonomous robots are expected to 
operate complex and complicated tasks 
such as autonomously driving in dynamic urban streets \cite{Herman_15}
or providing personalized services to clients \cite{kim2015socially}.
As an explicit rule-based programming has weakness in handling
diverse and exceptional uncommon cases, 
reinforcement learning (RL) has received a great deal of attention
\cite{kober2013reinforcement}. 
%The objective of RL is to find an optimal policy which maximizes 
%the expected sum or average of rewards,
%where the reward function defines an one step performance measure. 
% Consequently, 
When applying RL to practical real-world problems, 
designing an appropriate reward function 
is the most important part 
as it directly affects the quality of the resulting policy function. 

% 2. RL is good, but not always 
When it comes to problems for which explicitly 
designing a reward function is challenging, 
RL might not be an opportune choice to select.  
For example, 
manually describing a proper reward function that can generate
socially acceptable driving behavior styles \cite{Herman_15}
or acrobatic fight of a helicopter \cite{abbeel2010autonomous}
can be cumbersome. 
On the contrary, as humans can easily \textit{demonstrate} 
such behaviors, it is more effective to first collect  
demonstrations of an expert and let robots mimic the expert. 
This problem of imitating human-controlled behaviors is often 
referred to as imitation learning \cite{ross2010efficient}. 

% 3. What is an imitation learning 
Imitation learning can be categorized into two groups based on 
which function we are inferring about \cite{ross2010efficient}.
The first group focuses on directly mimicking the policy function 
which maps the state space or observations to the set of possible
actions. 
The other group focuses on reconstructing the underlying 
reward function, which is a succinct representation 
of an expert's objective \cite{ng2000algorithms}.
%in the mathematical framework of a  
%Markov decision process (MDP). 
This approach is often referred to as inverse optimal control 
\cite{Levine_12}
or inverse reinforcement learning (IRL) \cite{ng2000algorithms}. 
% 4. Advantage of IRL compared to DPL 
While it is often more difficult to infer the underlying reward function, 
IRL has a clear advantage over direct policy learning
in terms of generalization performance and transferability. 
In other words, the trained reward function can further be used
to obtain optimal control in unseen situations (generalization) 
or even to different model dynamics (transferability). 
% 5. Disadvantages of IRL 
However, most of the IRL methods require 
discretization of the state space and 
full specification of the model dynamics
as it solves a Markov decision problem as a subroutine. 
This obstructs model-based IRL methods to be applied 
to large scale systems with continuous state and action spaces. 
To overcome this issue, 
few model-free IRL methods have been proposed recently. 
However, these methods are often based on sampling to compute the 
reward function, which takes a fair amount of time. 

% 6. What we propose ? 
In this paper, we propose a density based
model-free inverse reinforcement learning 
algorithm, named density matching reward learning (DMRL). 
DMRL first computes the empirical distribution of 
expert's trajectories and finds a reward function that 
maximizes the inner-product between the empirical distribution
and the reward function. 
%inspired from the average reward reinforcement learning setting.
In other words, we optimize the reward function 
that best \textit{matches} the empirical state-action distribution of
an expert \cite{ho2016generative}. 
Furthermore, we perform theoretical analyses of the proposed method and 
derived the sample complexity of the proposed method under mild assumptions. 
To handle nonlinear reward functions, kernel DMRL (KDMRL) is also proposed. 
% 7. Experimental results
KDMRL has been applied to two experiments: grid world and track driving experiments. 
In grid world experiments, KDMRL shows superior performance for cases
with nonlinear reward functions and comparable performance to linear
IRL methods for linear reward functions in terms of expected value differences.
In track driving experiments, KDMRL outperforms all other compared methods
in terms of mimicking experts' driving style as well as 
safety and stability of driving. 

%%%%%%%%%%%%%%%%%%%%%%%%%%%%%%%%
% 									Related Works
%%%%%%%%%%%%%%%%%%%%%%%%%%%%%%%%
\section{Related Work} \label{sec:rel}

Recently proposed inverse reinforcement learning (IRL) 
methods can be grouped by two different criteria. 
The first criterion is about how we define the objective function 
of the IRL problem: 
margin based or probabilistic model based methods. 
On the other hand, the second criterion is about the necessity of 
the transition model, i.e., $P(s' | s, a)$, of the underlying 
Markov decision process. 
The classification of state-of-the-art IRL algorithms, including ours,
is summarized in Table \ref{tbl:svy}. 

\begin{table}[t]
\centering       
\resizebox{0.7\textwidth}{!}{
\begin{tabular}{|l|l|l|l|l|}
\hline
&{Model-based} &{Model-free} \\ \hline
Margin based model
	&  
	\cite{abbeel2004apprenticeship, 
	ratliff2006maximum, ratliff2009learning} 
	&    
	\cite{Klein2012scirl}
\\ \hline
Probabilistic model 
	&   
	\cite{ziebart2008maximum, dvijotham2010inverse, 
	levine2011nonlinear,
	wulfmeier2015maximum, shiarlis2015inverse}
	&  
	\cite{boularias2011relative, Finn2016gcl, Herman2016inverse}, Ours                  
\\ \hline
\end{tabular}
}

\caption{Classification of IRL algorithms}
\label{tbl:svy}
\end{table} 
%% Margin-based method 
A margin based approach aims to maximize the margin 
between the value of the expert's policy compared to other possible policies. 
An IRL problem was first introduced by Ng and Russell
\cite{ng2000algorithms},
where the parameters of the reward function are optimized by 
maximizing the margin between randomly sampled reward functions. 
Abbeel and Ng \cite{abbeel2004apprenticeship} proposed an
apprenticeship learning algorithm,
which minimizes the difference  between the empirical distribution
of an expert and the distribution induced by the reward function.  
Ratliff et al. \cite{ratliff2006maximum} proposed the maximum
margin planning (MMP) algorithm, where Bellman-flow constraints are
utilized to maximize the margin between the experts' policy and all other
policies. 
%In \cite{ratliff2007boosting} and \cite{ratliff2009learning},
%Ratliff et al. extended the MMP by boosting and allowing
%learning a nonlinear reward function and proposed 
%MMP with boosting and 
%learning to search, respectively. 

On the other hand, probabilistic model based methods 
aim to find the reward function which maximizes the likelihood 
or posterior distribution. % of an expert's demonstration. 
%To define the probability distribution of expert's demonstrations, 
%a number of methods utilize a stochastic policy. 
%A stochastic policy model was proposed in 
%\cite{ziebart2008maximum, ramachandran2007bayesian}, where the authors 
%try to handle uncertainty or inconsistency of expert's behaviors.
Zeibart et al. \cite{ziebart2008maximum} proposed 
maximum entropy IRL (MaxEnt) using the principle of 
maximum entropy to alleviate the inherent ambiguity problem of 
the reward function.
%Furthermore, an efficient method to compute the gradient of 
%the likelihood function was also proposed.
In \cite{ramachandran2007bayesian}, Ramachandran et al. 
proposed Bayesian IRL (BIRL), where the Bayesian probabilistic model 
was defined and effectively solved by using 
a Metropolis-Hastings sampling method. 
%% where various prior distributions were also discussed. 
Levine and Koltun \cite{levine2011nonlinear} proposed Gaussian
process inverse reinforcement learning (GPIRL),
where a nonlinear reward function is effectively modeled 
by a sparse Gaussian process with inducing inputs. 

%Robust Bayesian inverse reinforcement learning (RBIRL) was proposed in
%\cite{zheng2014robust}, which is an extension of BIRL to handle
%noisy demonstrations. 
%RBIRL can automatically identify and remove noises in demonstrations
%using the expectation maximization algorithm.
%Wulfmeier et al. \cite{wulfmeier2015maximum} proposed maximum entropy
%deep inverse reinforcement learning, where the reward
%function is modeled by a neural network and is learned by maximizing
%the log likelihood of demonstrations using the method from
%\cite{ziebart2008maximum}. 

% Model-free 
%While a number of IRL methods have been effectively tackled 
%many practical problems such as acrobatic helicopter 
%maneuver \cite{abbeel2004apprenticeship}, 
%there remains a problem of solving a Markov decision process
%as its subroutine. 
%To handle this issue, several model-free IRL methods have been 
%recently proposed. 
Recently, several model-free IRL methods have been 
proposed. 
In \cite{boularias2011relative}, Boularias et al. proposed 
a relative entropy IRL (RelEnt) method, 
which minimizes the relative entropy between 
the empirical distribution of demonstrations from the 
baseline policy and the distribution under the learned policy.
Structured classification based IRL \cite{Klein2012scirl}
was proposed by formulating IRL problem as a 
structured multi-class classification problem
by founding that the parameters of the action-value function 
are shared with those of the reward function.
%In particular, they found that the parameters of the action-value function
%are shared with the parameters of the reward function
%and this property is utilized to formulate a structured
%classification problem.  
In \cite{Herman2016inverse}, Herman et al. proposed a
model-free IRL method by simultaneously estimating the reward 
and dynamics, where they separate the real transition model (dynamics)
and the belief transition model of an agent to handle previously
unseen states.
Finn et al. \cite{Finn2016gcl} proposed a guided cost learning, 
which extends the maximum entropy IRL to a model-free setting based on
approximation by importance sampling. 
{In this paper, we propose a model-free IRL method
that first compute the empirical state action distribution of an expert
and then find the reward function that matches the density function. 
}
%The proposal distribution for importance sampling is derived from 
%policy optimization with differential dynamic programming. 

%{\bf [It will be useful to discuss how the proposed method is compared to existing methods.]}

%%%%%%%%%%%%%%%%%%%%%%%%%%%%%%%%
% 									Proposed Algorithm
%%%%%%%%%%%%%%%%%%%%%%%%%%%%%%%%
\section{Problem Formulation}

Before presenting the proposed inverse reinforcement learning (IRL) method, 
let us introduce the objective of reinforcement learning (RL). 
In particular, we will consider an average reward RL problem, which
often has advantages over discounted formulations with respect to
stabilities as we do not have to choose a discount factor 
or time horizon \cite{kober2013reinforcement}.
An average reward RL problem can be
formulated as follows:  
\begin{equation}
\begin{aligned} \label{eqn:rl1}
	\underset{\pi}{\text{maximize}} 
	&& V(\pi) & = \sum_{s,a}\mu^{\pi}(s)\pi(s,a)R(s,a) \\
	\textrm{subject to} 
	&& \mu^{\pi}(s') & = \sum_{s,a} \mu^{\pi}(s)\pi(s,a)T(s,a,s'), \forall s' \in S \\
	&& 1 & = \sum_{s,a} \mu^{\pi}(s)\pi(s,a) \\
	&& \pi(s,a) & \ge 0, \forall s \in S, a \in A, \\
\end{aligned}
\end{equation}
where 
$\pi(s, a)$ is a conditional distribution of action $a$ given state $s$, i.e., 
$\pi(s, a) = p(a | s)$, and 
$\mu^{\pi}(s)$ is a stationary state distribution induced by 
policy $\pi$. 
Using a chain rule, we can easily define a joint state-action distribution 
$\mu(s, a) = \mu^{\pi}(s)\pi(s, a)$
and rewrite the RL formulation in (\ref{eqn:rl1}) as follows:
\begin{equation}
\begin{aligned} \label{eqn:rl2}
	\underset{\mu}{\text{maximize}} 
	&& V(\mu) & = \langle \mu, R \rangle \\
	\textrm{subject to} 
	&& \mu &\in G(\mu), \\
\end{aligned}
\end{equation}
where $\langle \cdot, \cdot \rangle$
indicates an inner-product operation and
$G(\mu)$ is a feasible set of $\mu$
induced by the model dynamics
which is often referred to as a Bellman flow constraint. 
Intuitively speaking, the objective of RL in (\ref{eqn:rl2}) 
is to find a stationary state-action distribution $\mu(s, a)$
given a reward function $R(s, a)$ considering a Bellman flow
constraint $G(\mu)$,
where the policy function $\pi(s,a)$ can naturally be
derived from $\mu(s, a)$ as
$\pi(s, a) = \frac{\mu(s, a)}{\sum_a \mu(s, a)} $\cite{kober2013reinforcement}.

As the objective of IRL is to 
recover $R(s, a)$ given a set of
$N$ trajectories $\Xi = \{ \xi_i \}_i^{N}$,
where a single trajectory $\xi_i$ consists of $M$ state-action pairs, 
$\xi_i =(s_{ij}, a_{ij})_{j=1}^M$,
existing IRL methods 
either try to maximize a margin between 
values of expert's trajectories and other sampled
or every possible values 
\cite{ng2000algorithms, abbeel2004apprenticeship, ratliff2006maximum}
or estimate the parameters of a reward function
by properly defining the likelihood (probability density)
of trajectories $\Xi$
\cite{ziebart2008maximum, levine2011nonlinear, ramachandran2007bayesian, boularias2011relative, wulfmeier2015maximum}.

% What we do! 
In this paper, we cast the reward estimation problem as a 
density matching problem. 
In particular, 
we first assume that the demonstrations are sampled from 
the stationary state-action distribution $\mu(s, a)$
and estimate the probability density of the state-action space
using a set of expert's trajectories $\Xi$. 
In this paper, kernel density estimation with radial basis functions
\cite{Zeevi_97} is used to estimate the joint state-action density as
$\hat{\mu}(s, a)$, but other suitable estimator can be applied.
Then, the underlying reward function is estimated from the 
following optimization problem, which we call 
density matching reward learning (DMRL): 
\begin{equation}
\begin{aligned} \label{eqn:dmrl}
	\underset{R}{\text{maximize}} 
	&& V(R) & = \langle \hat{\mu}, R \rangle \\
	\textrm{subject to} 
	&& \| R \|_2 & \le 1, \\
\end{aligned}
\end{equation}
where the norm ball constraint $\| R \|_2 \le 1$ 
is introduced to handle
the scale ambiguity of the 
reward function \cite{ng2000algorithms} and $\langle \hat{\mu}, R \rangle 
= \int_{\mathcal{S} \times \mathcal{A}} \hat{\mu}(s,a)R(s,a) \, ds \, da$. 

Suppose that we have access to the true stationary state-action density
$\bar{\mu}(s, a)$ and estimate the reward $\bar{R}(s, a)$ from (\ref{eqn:dmrl}),
i.e., $\bar{R} = \arg \max_{ \norm{R}_2 \le 1} \langle \bar{\mu}, R \rangle $.
We define the value of $\bar{R}$ with respect to the true state-action density
$\bar{\mu}(s, a)$ as an optimal value $\bar{V} = \langle \bar{\mu}, \bar{R} \rangle$.
In other words, the optimal value $\bar{V}$ is an expectation of reward $\bar{R}$
with respect to the true state-action density $\bar{\mu}$, i.e.,
\begin{equation} \label{eqn:optv}
	\bar{V} = \langle \bar{\mu}, \bar{R} \rangle.
\end{equation}

Similarly, we define an estimated value $\hat{V}$
as an expectation of an estimated reward $\hat{R}$
computed from solving  (\ref{eqn:dmrl}) with estimated density $\hat{\mu}$
with respect to the true state action density
$\bar{\mu}(s, a)$, i.e.,
\begin{equation} \label{eqn:estv}
	\hat{V} = \langle \bar{\mu}, \hat{R} \rangle.
\end{equation}
% {\bf [Need to justify the use of $\hat{\mu}$ here.]}

%%%%%%%%%% Definition: Optimal Value
%\begin{definition}\label{def:optv}
%	Optimal value $\bar{V}$ is a conditional expectation of estimated 
%	reward $\bar{R}$ from the true state-action density $\mu$, i.e., 
%	$\bar{V} = \langle \mu, \bar{R} \rangle$.
%\end{definition}

%%%%%%%%%% Definition: Estimated Value
%\begin{definition}\label{def:estv}
%	Estimated value $\hat{V}$ is a conditional expectation of estimated 
%	reward $\bar{R}$ from estimated state-action density $\hat{\mu}$
%	from expert's trajectory $\xi$, i.e., 
%	$\bar{V} = \langle \mu, \bar{R} \rangle$.
%\end{definition}

Of course, estimating the true density $\bar{\mu}(s, a)$ is practically infeasible 
as it requires an infinite number of samples. 
In Section \ref{sec:the}, 
we give a probabilistic bound on the absolute difference between $\bar{V}$
and $\hat{V}$, i.e., $| \bar{V} - \hat{V} |$
as a function of the number of trajectories (sample complexity).
Furthermore, the proposed method can be easily extended to a continuous 
domain using the kernel method \cite{quinonero2005unifying} as
described in Section \ref{sec:kdmrl}.

%%%%%%%%%%%%%%%%%%%%%%%%%%%%%%%
% 									Theoretical Analysis
%%%%%%%%%%%%%%%%%%%%%%%%%%%%%%%%
\section{Theoretical Analysis} \label{sec:the}

In this section, we present a bound on the sample complexity of the
proposed method, DMRL.  
In particular, we will focus on the absolute difference between the
optimal value $\bar{V}$ and estimated value $\hat{V}$,
i.e., $| \bar{V} - \hat{V} |$. 
We would like to note that this analysis is not based on  
the expected value difference (EVD), which is used in
\cite{abbeel2004apprenticeship, Klein2012scirl} 
to theoretically analyze the performance of IRL methods. 
This is mainly because only reward functions linear with respect to
parameters can be analyzed using EVDs. 
We instead show the sample complexity on the absolute difference
between the best value $\bar{V}$ 
and the optimized value, $\hat{V}$, with $n$ samples. 
As $| \bar{V} - \hat{V} | = | \langle \bar{\mu}, \, \bar{R}-\hat{R} \rangle |$, 
minimizing this quantity can be interpreted as minimizing the
difference between the estimated reward $\hat{R}$ and the true reward
$\bar{R}$ when projected by the true state-action density $\bar{\mu}$.

Followings are the main results of this paper. 
%%%%%%%%%% Main Theorem 1
\begin {theorem} \label{thm:main1}
	Let $\hat{R}$ be computed from 
	$\hat{R} = \underset{R}{\mathrm{arg\max} \langle \hat{\mu}, R \rangle} \ \text{s.t.} \ \| R \|_2 \le 1$
	and $\bar{R}$ be computed from 
	$\bar{R} = \underset{R}{\mathrm{arg\max} \langle \bar{\mu}, R \rangle} \ \text{s.t.} \ \| R \|_2 \le 1$,
	where $\hat{\mu}$ and $\bar{\mu}$ are estimated and true
	densities, respectively. 
	Then
	\begin{equation*}
		| \langle \bar{\mu}, \hat{R} \rangle - \langle \bar{\mu}, \bar{R} \rangle | 
		\le
		3(R_{max} - R_{min})d_{var}(\bar{\mu}, \hat{\mu})
	\end{equation*}
	where $d_{var}(\cdot, \cdot)$ is the variational distance\footnote{
		The variational distance between two probability
		distributions, $P$ and $Q$, is defined as
		$d_{var}(P, Q) = \frac{1}{2}\sum_{x \in \mathcal{X}}|P(x)-Q(x)|$.
		It is also known that 
		$d_{var}(P, Q) = \sum_{x \in \mathcal{X}:P(x)>Q(x)}|P(x)-Q(x)|$.
		}
		between two probability distributions. 
\end {theorem}

%%%%%%%%%% Main Theorem 2
\begin {theorem} \label{thm:main2}
	Suppose Assumption \ref{ass:1}--\ref{ass:8} in Appendix~\ref{appendix} hold.
	Let $n$ and $N$ be the number of samples and the number of
	basis functions for kernel density esimation, respectively.
%%	and $\hat{f}_{n, N}$ be the estimated density function 
%% 	from kernel density estimation with $n$ samples and $N$ basis functions. 
	Then, for any $\delta \ge 0$ and $n$ and $N$ be sufficiently
	large, we have
	\begin{equation}
		|\langle \bar{\mu}, \hat{R} \rangle - \langle \bar{\mu}, \bar{R} \rangle| 
		\le 3\sqrt{2} (R_{max}-R_{min}) \sqrt{ O\left( \frac{1}{n} \right) 
			+ O\left( \frac{nd}{N}(1+\frac{1}{\sqrt{\delta}}) \right) }
	\end{equation}
	with probability at least $(1-\delta)$.
\end {theorem}

The main intuition behind the proposed DMRL is that IRL can be viewed
as a dual problem of finding a reward function that matches 
the state-action distribution \cite{ho2016generative}. 
The domain of our interest is the product space of 
states and actions, $\mathcal{S} \times \mathcal{A}$,
and for the notational simplicity, we will denote this space as 
$\mathcal{X} = \mathcal{S} \times \mathcal{A}$.
Before proving Theorem \ref{thm:main1} and \ref{thm:main2},
let us first introduce useful lemmas. 

%%%%%%%%%% Lemma 1
\begin{lemma} {\cite{gilks2005markov}}
	\label{lem:1}
	Let $P$ and $Q$ be probability distributions over $\mathcal{X}$
	and $f$ be a bounded function on $\mathcal{X}$.
	Then,
	\begin{equation*}
		|\langle P, f \rangle - \langle Q, f \rangle| \le (\sup f - \inf f)d_{var}(P, Q).
	\end{equation*}
\end{lemma}
%%%%%%%%%% Proof of Lemma 1
%\begin{proof}
%	Proof can be found in \cite{abbeel2004apprenticeship}.
%	\begin{equation*}
%		\begin{aligned}
%			&& \langle P, \, f \rangle - \langle Q, \, f \rangle
%			&      = \langle P-Q, \, f - \inf f \rangle + \langle P-Q, \, \inf f \rangle \\
%			&& & = \sum_{x: P(x) > Q(x)} (f(x)-\inf f)(P(x)-Q(x)) \\
%			&& &	  \quad  \quad \quad + \sum_{x: P(x) \le Q(x)} (f(x)-\inf f)(P(x)-Q(x)) \\
%			&& & \le  \sum_{x: P(x) > Q(x)} (f(x)-\inf f)(P(x)-Q(x)) \\
%			&& & \le   (\sup f - \inf f)d_{var}(P, Q)
%		\end{aligned}
%	\end{equation*}
%	Since $d_{var}(P, \, Q) = d_{var}(Q, \, P)$
%	this completes the proof. 
%\end{proof}

%%%%%%%%%% Lemma 2
\begin{lemma} {\cite{Pollard_00}}
	\label{lem:2}
	Suppose $P$ and $Q$ are probability distributions. 
	Then, 
	\begin{equation*}
		d_{var}(P, \, Q) \le \sqrt{2} d_{\mathcal{H}}(P, \, Q),
	\end{equation*}
	where $d_{\mathcal{H}}(\cdot)$ is the Hellinger distance\footnote{
		The Hellinger distance between two probability distributions $P$ and $Q$
		is defined as $d^2_{\mathcal{H}}(P, Q) = \frac{1}{2}
			\sum_{x \in \mathcal{X}} (\sqrt{P(x)} - \sqrt{Q(x)})^2
		$
	}.
\end{lemma}
%%%%%%%%%% Proof of Lemma 2
%\begin{proof}
%	Proof can be found in \cite{Pollard_00}.
%	\begin{equation*}
%		\begin{aligned}
%			&& |d_{var}(P, Q)|^2 
%			& = \frac{1}{4} \left( \sum_{x \in \mathcal{X}} |P(x)-Q(x)| \right)^2 \\
%			&& & = \frac{1}{4} \left( \sum_{x\in\mathcal{X}} (\sqrt{P(x)}-\sqrt{Q(x)})(\sqrt{P(x)}
%			+\sqrt{Q(x)}) \right)^2 \\
%			&& & \le \frac{1}{4} \left( \sum_{x \in \mathcal{X}} (\sqrt{P(x)}-\sqrt{Q(x)})^2 \right)
%				\left( \sum_{x \in \mathcal{X}} (\sqrt{P(x)}+\sqrt{Q(x)})^2 \right) \\
%			&& & = \frac{1}{2} d^2_H(P, Q) \left(2 + 2 \sum_{x \in \mathcal{X}} \sqrt{P(x)Q(x)} \right)	\\
%			&& & = d^2_H(P, Q) \left( 2 - d^2_H(P, Q) \right) \\
%			&& & \le 2 d^2_H(P, Q)
%		\end{aligned}
%	\end{equation*}		
%\end{proof}

%%%%%%%%%% Lemma 3
\begin{lemma}{\cite{Alfeld_04}}\label{lem:3} 
	Suppose $\| \cdot \|$ is a proper norm defined on index set $\mathcal{X}$.
	Then, for any $a, b \in \mathcal{X} $,
	\begin{equation*}
		|\| a \| - \| b \|| \le \| a-b \|.
	\end{equation*}
\end{lemma}
%%%%%%%%%% Proof of Lemma 3
%\begin{proof}
%	Proof can be found in \cite{Alfeld_04}.
%	The Lemma \ref{lem:3} is often referred to as a reverse triangle inequality. 
%	As a proper norm function satisfies triangle inequality, following two equations are correct:
%	\begin{equation}\label{eq:te1}
%		\| a \| + \| b - a \| \ge \| b \|
%	\end{equation}
%	\begin{equation}\label{eq:te2}
%		\| b \| + \| a - b \| \ge \| a \|
%	\end{equation}
%	Rearranging (\ref{eq:te1}) and (\ref{eq:te2}), we get
%	\begin{equation}\label{eq:te3}
%		\| b - a \| \ge \| b \| - \| a \| 
%	\end{equation}
%	\begin{equation}\label{eq:te4}
%		\| a - b \| \ge \| a \| - \| b \|
%	\end{equation}
%	As we $\| a - b \| = \| b - a \|$, combining (\ref{eq:te3}) and (\ref{eq:te4}),
%	we get the reverse triangle inequality.
%\end{proof}

We also introduce a theorem from \cite{Zeevi_97}, which is used to
prove Theorem \ref{thm:main2}. 
%%%%%%%%%% Theorem [Zeevi1997]
\begin{theorem} \label{thm:Zeevi_97}
	Suppose Assumption \ref{ass:1}--\ref{ass:8} in
	Appendix~\ref{appendix} hold.
	Let $n$ and $N$ be the number of samples and basis functions for
	kernel density esimation, respectively, and $\hat{f}_{n, N}$
	be the estimated density function from kernel density estimation 
	with $n$ samples and $N$ basis functions. 	
	Then, for any $\delta \ge 0$ and $n$ and $N$ sufficiently large,
	we have
	\begin{equation*}
		d^2_{\mathcal{H}}(f, \hat{f}_{n,N}) 
			\le O\left( \frac{1}{n} \right) 
			+ O\left( \frac{nd}{N}(1+\frac{1}{\sqrt{\delta}}) \right),
	\end{equation*}
	where $d^2_{\mathcal{H}}(\cdot)$ is a squared Hellinger distance between
	two probability distributions. 
\end{theorem}

We are now ready to prove Theorem \ref{thm:main1} and Theorem \ref{thm:main2}.

%%%%%%%%%% Proof of Theorem1
\begin{proof}[Theorem \ref{thm:main1}]
	Let $\bar{\mu}$ and $\hat{\mu}$ be the true and estimated density functions, 
	respectively, and $\bar{R}$ and $\hat{R}$ be the reward functions
	estimated by DMRL with $\bar{\mu}$ and $\hat{\mu}$, respectively. 
	The absolute difference between the optimal value $\bar{V}$ and 
	the estimated value $\hat{V}$ can be expressed as:
	\begin{equation*}
		\begin{aligned} 
		&& |\langle \bar{\mu}, \hat{R} \rangle - \langle \bar{\mu}, \bar{R} \rangle| 
			& = |\langle \bar{\mu}, \hat{R} \rangle - \langle \bar{\mu}, \bar{R} \rangle
				+ \langle \hat{\mu}, \hat{R} \rangle - \langle \hat{\mu}, \hat{R} \rangle| \\
		&&  & \le |\langle \bar{\mu}, \hat{R} \rangle - \langle \hat{\mu}, \hat{R} \rangle |
				+ |\langle \bar{\mu}, \bar{R} \rangle - \langle \hat{\mu}, \hat{R} \rangle| \\
		&&  & \le (R_{max}-R_{min}) d_{var}(\bar{\mu}, \, \hat{\mu}) 
			+ | \langle \bar{\mu}, \, \bar{R} \rangle - \langle \hat{\mu}, \, \hat{R} \rangle |,
		\end{aligned}
	\end{equation*}
	where we used the triangular inequality and Lemma \ref{lem:1}.

	$| \langle \bar{\mu}, \, \bar{R} \rangle - \langle \hat{\mu}, \, \hat{R} \rangle |$
	can be further bounded by 
	\begin{equation*}
		\begin{aligned}
		&& |\langle \bar{\mu}, \bar{R} \rangle - \langle \hat{\mu}, \hat{R} \rangle | 
			& = \left|  \max_{\| R\| \le 1}{\langle \bar{\mu}, R \rangle} 
			- \max_{\| R\| \le 1}{\langle \hat{\mu}, R \rangle} \right| \\
		&&  & \le  \max_{\| R\| \le 1}{\langle \bar{\mu} - \hat{\mu}, R \rangle}  \\
		&&  & \le  (R_{max}-R_{min}) \sum_{x \in \mathcal{X}} |\mu(x) - \hat{\mu}(x)|  \\
		&&  & =   2(R_{max}-R_{min}) d_{var}(\bar{\mu}, \hat{\mu}),
		\end{aligned}
	\end{equation*}
	where we used the definition of optimal and estimated values, 
	Lemma \ref{lem:3} for a dual norm, the definition of an inner product, 
	and the definition of $d_{var}(\cdot)$. 
\end{proof}
%%%%%%%%%% Proof of Theorem2
\begin{proof}[Theorem \ref{thm:main2}]
	Combining aforementioned two inequalities and Lemma \ref{lem:2},
	we get
	\begin{equation*}
		\begin{aligned}
		&& |\langle \bar{\mu}, \hat{R} \rangle - \langle \bar{\mu}, \bar{R} \rangle| 
		 &\le 3(R_{max}-R_{min})d_{var}(\bar{\mu}, \, \hat{\mu}) \\
		&&  &\le 3\sqrt{2} (R_{max}-R_{min}) \sqrt{ O\left( \frac{1}{n} \right) 
			+ O\left( \frac{nd}{N}(1+\frac{1}{\sqrt{\delta}}) \right) }.
		\end{aligned}
	\end{equation*}
\end{proof}

%%%%%%%%%%%%%%%%%%%%%%%%%%%%%%%%
% 									
%%%%%%%%%%%%%%%%%%%%%%%%%%%%%%%%
\section{Kernel DMRL} \label{sec:kdmrl}

The proposed DMRL method can be extended to a continuous state-action space 
using a kernel-based function approximation method,
which will be referred to as kernel density matching 
reward learning (KDMRL). 
In particular, we assume that the reward function is in the 
reproducing kernel Hilbert space (RKHS) with a finite set of inducing inputs.
The original DMRL formulation in (\ref{eqn:dmrl})
can be relaxed by solving the following optimization problem:
\begin{equation}
\begin{aligned} \label{eqn:dmrl2}
	\underset{\tilde{R}}{\text{maximize}} 
	&&  \tilde{V} = \sum_{\forall x \, \in U} \hat{\mu}(x)\tilde{R}(x) 
		- \frac{\lambda}{2} \| \tilde{R} \|_{\mathcal{H}}^2, \\
\end{aligned}
\end{equation}
where 
$\hat{\mu}$ is the estimated density function, 
$\lambda$ controls the smoothness of the reward function, 
$U = \{ x^U_i \}_{i=1}^{N_U}$ is a set 
of $N_U$ inducing points, and
$\| \tilde{R} \|_{\mathcal{H}}^2$ is the squared Hilbert norm,
which is often used as a regularizer for kernel machines.
Then, the reward function with inducing inputs can be expressed as 
\begin{equation}
	\tilde{R}(x) = 
	\sum_{i=1}^{N_U} \alpha_i k(x, \, {x^U_{i}}),
\end{equation}
where $x \in \mathcal{X}$ is a state-action pair,
$U \subset \mathcal{X} $ is a set of pre-defined inducing points,
$\alpha \in \R^{N_U}$ is a parameter determining the shape 
of the reward function, and $k(\cdot, \cdot)$ is a positive semidefinite kernel function. 
In other words, optimizing $\tilde{R}$ in (\ref{eqn:dmrl2}) is equivalent to 
optimizing $N_u$ dimensional vector $\alpha$ and 
the squared Hilbert norm of $\tilde{R}$, where 
$\norm{\tilde{R}}_{\mathcal{H}}^2 = \alpha^T K_{\mathbf{u}}\alpha$
and $K_{\mathbf{u}}$ is a kernel matrix computed from $U$.

We use kernel density estimation (KDE) to estimate the
probability density function 
\begin{equation} \label{eqn:kde}
	\hat{\mu}(x) = \frac{1}{N_D}\sum_{k=1}^{N_D} k_{\mu}(x, \, x^D_k),
\end{equation}
where $N_D$ is the number of training samples, $x^D_k$ is the $k$th
training data, and $k_{\mu}(\cdot, x^D)$ is a kernel function 
whose integral is one (basis density function). 
Then, (\ref{eqn:dmrl2}) can be reformulated as:
\begin{equation}
\begin{aligned} \label{eqn:dmrl3}
	\underset{\alpha}{\text{max}} 
	&&  \tilde{V} & = \sum_{i=1}^{N_U}  \sum_{j=1}^{N_U} 
		\hat{\mu}(x^U_i)  \alpha_j k(x^U_i, \, x^U_j) 
		- \frac{\lambda}{2} \alpha^T K_{U}\alpha \\
	&&  & = 
		\frac{1}{N_D}\sum_{i=1}^{N_U}  \sum_{j=1}^{N_U} \sum_{k=1}^{N_D}
		% \hat{\mu}(x^U_i)  
		k_{\mu}(x_i^U, \, x^D_k)
		\alpha_j k(x^U_i, \, x^U_j)
		- \frac{\lambda}{2} \alpha^T K_{U}\alpha, \\
\end{aligned}
\end{equation}
which can be easily solved by quadratic programming.

% Leverage..
The objective of the kernel density estimation in (\ref{eqn:kde})
is to estimate the stationary distribution of an MDP. 
If we have infinite length trajectories, $\hat{\mu}$ converges
to the true $\bar{\mu}$ regardless of the initial state
distribution by the Ergodic theorem \cite{gilks2005markov}\footnote{
Here, we assume that the state transition 
probability induced by the expert's policy is ergodic.}. 
However, as we have finite length trajectories, we have 
to compensate the effect of the initial distribution. 
In Markov chain Monte Carlo (MCMC), some initial samples known as
burn-in samples are ignored when estimates are computed. 
However, in our case, as the length of a demonstration is not long
enough to reserve burn-in samples, we differentiate the influence of
each sample by giving more weights to the latter ones. 
In this regards, we modify the density function 
(\ref{eqn:kde}) using a leveraged kernel function 
proposed in \cite{choi2016smooth}
\begin{equation} \label{eqn:kde2}
	\hat{\mu}(x) = \frac{1}{Z}\sum_{k=1}^{N_D} 
		\cos\left(\frac{\pi}{2}(1-\gamma_k)\right) k_{\mu}(x, x^D_k),
\end{equation}
where $\gamma_k = \delta^{T-t}$ indicates the leverage (or influence)
of the $k$th training data, $0 < \delta \le 1$ is a constant, 
$T$ and $t$ are the length and time-step of the trajectory, 
and $Z$ is a normalization constant. 
%In other words, $\gamma$ represents the importance of each data. 
%In our experiments, we define $\gamma = \rho^{T-t}$ 
%where $\rho$ is a discount factor whose value is between $0$ and $1$
%and $t$ is a discrete time index and $T$ is the length of a trajectory. 
%Intuitively speaking, we give more weights to the latter ones. 
We can rewrite (\ref{eqn:dmrl3}) with an additional two norm regularization,
$\frac{\beta}{2}\norm{\alpha}_2^2$,
in a matrix form as follows:
\begin{equation}
\begin{aligned} \label{eqn:dmrl4}
	\underset{\alpha}{\text{max}} 
	&&  \tilde{V} = \alpha^T K_{U} K_D \mathbf{1}_{N_D}
		- \frac{\lambda}{2} \alpha^T K_{U}\alpha - \frac{\beta}{2}\alpha^T \alpha, \\
\end{aligned}
\end{equation}
%where $\mu_{\mathbf{u}} \in \R_{\ge 0}^{N_D}$ 
%is a vector whose $i$th element is a probability density computed 
%at $i$th inducing point and $[\mathbf{x}]_i$ is $i$th element of 
%a vector $\mathbf{x}$. 
where $[{K_U}]_{ij} = k(x^U_i, x^U_j)$,
$[{K_D}]_{ij} = k_{\mu}(x^U_i, x^D_j)$,
$\lambda$ and $\beta$ are regularization coefficients,
and $\mathbf{1}_{N_D} \in \R^{N_D}$ is a vector of $1$'s. 
As (\ref{eqn:dmrl4}) is quadratic with respect to the
optimization variable $\alpha$, (\ref{eqn:dmrl4}) has an analytic
solution: 
\begin{equation}
	\hat{\alpha} = \frac{1}{N_D}(\beta K_U + \lambda I)^{-1}K_U K_D \mathbf{1}_{N_D}.
\end{equation}
In following experiments, we use a widely used squared exponential 
kernel function for both $k(\cdot, \, \cdot)$ and $k_{\mu}(\cdot, \, \cdot)$. 
While it is possible to optimize the hyperparameters of the kernel function 
by maximizing (\ref{eqn:dmrl4}), we use a simple median trick \cite{Dai_14}
for determining the hyperparameters. 

%%%%%%%%%%%%%%%%%%%%%%%%%%%%%%%%
% 									
%%%%%%%%%%%%%%%%%%%%%%%%%%%%%%%%
%\subsection{Incorporating Negative Demonstrations}
%
%It is often the case that the objective of a robot is not only described 
%by what it should do, but also by what it should not do. 
%In such cases, we can generate both positive and negative 
%demonstrations and compute two different 
%density functions, $\hat{\mu}^+(s)$ and $\hat{\mu}^-(s)$,
%from two different sets of demonstrations.  
%
%Using $\hat{\mu}^+(s)$ and $\hat{\mu}^-(s)$, we can 
%naturally extend (\ref{eqn:dmrl4}) to following reward learning problem 
%as follows:
%\begin{equation}
%\begin{aligned} \label{eqn:dmrl5}
%	\underset{\alpha}{\text{maximize}} 
%	&&  \tilde{V} = \alpha^T K_{\mathbf{u}} \mu_{\mathbf{u}}^+
%		- \alpha^T K_{\mathbf{u}} \mu_{\mathbf{u}}^-
%		- \lambda \alpha^T K_{\mathbf{u}}\alpha \\
%\end{aligned}
%\end{equation}
%where $\mu_{\mathbf{u}}^+$ and $\mu_{\mathbf{u}}^-$
%are density vectors computed from 
%$\hat{\mu}^+(s)$ and $\hat{\mu}^-(s)$, respectively. 

%%%%%%%%%%%%%%%%%%%%%%%%%%%%%%%%
% 									Experiments
%%%%%%%%%%%%%%%%%%%%%%%%%%%%%%%%
\section{Experiments} \label{sec:exp}

To validate the performance of the proposed KDMRL method,
two different scenarios have been investigated:
grid world and track driving experiments.
The grid world experiments are conducted for 
quantitatively comparing the performance of KDMRL against 
four different inverse reinforcement learning methods,
maximum margin planning (MMP) \cite{ratliff2006maximum},
relative entropy inverse reinforcement learning (RelEnt) 
\cite{boularias2011relative},
maximum entropy inverse reinforcement learning (MaxEnt) 
\cite{ziebart2008maximum}, and
Gaussian process inverse reinforcement learning (GPIRL) 
\cite{levine2011nonlinear}. 
In the track driving experiments, we first collect three different types
of driving styles by manually controlling a car in a simulated environment
inspired by \cite{Levine_12}. 
From the collected demonstrations, an underlying reward function
is inferred and used to control the car with receding horizon control (RHC).
% in both tracks used for collecting the
%demonstrations and more complex unseen tracks with moving cars. 
In this case, MaxEnt, RelEnt, and GPIRL have been compared 
with KDMRL. 

%%%%%%%%%%%%%%%%%%%%%%%%%%%%%%%%
% 									Grid World
%%%%%%%%%%%%%%%%%%%%%%%%%%%%%%%%
\subsection{Grid World Experiments}

%\begin{figure}[!t] \centering
%	\subfigure[]{\includegraphics[width=.37\columnwidth]{figures/fig_16LinTime} \label{fig:grid16lin_time}}
%	\subfigure[]{\includegraphics[width=.37\columnwidth]{figures/fig_32LinTime} \label{fig:grid32lin_time}}
%	\caption{
%		Computations times of compared IRL methods are shown
%		as a function of the number of trajectories.
%		(a) Small $16 \times 16$ grid world with linear reward function 
%		(b) Large $32 \times 32$ grid world with linear reward function 
%		}
%	\label{fig:gridtime}
%\end{figure}

% Grid world setting 
%% We first validate the performance of the proposed KDMRL 
%% on the grid world experiment where the objective is to find the reward function 
%% based on the demonstrations of experts.  
% who maximize the expected sum of rewards collected from the traversed reward field. 
In a grid world, an expert can make five actions, up, down, left, right, and staying. 
The underlying reward function $R(s)$ is defined as 
$R(s) = \sum_{i=1}^k \alpha_i \exp(-\norm{s-c_i} _2^2)$,
where $c_i$ is the $i$th randomly-deployed
location, $k$ is the number of peak points, and 
$\alpha_i$ is randomly sampled from $\{ -1, 1\}$.
In other words, the reward functions has $k$
randomly located concave or convex peaks.

\begin{figure}[!t] \centering
	\subfigure[]{\includegraphics[width=.42\columnwidth]{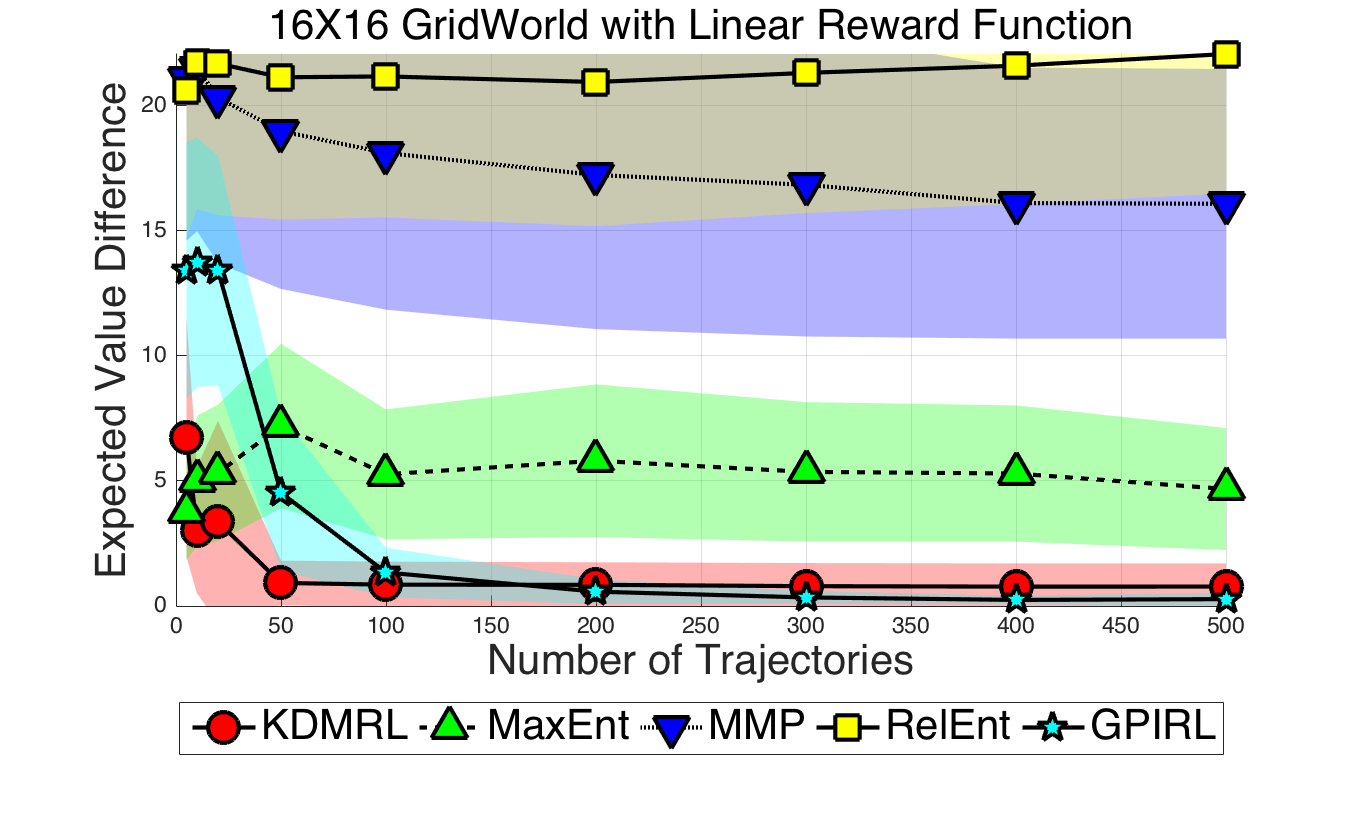} \label{fig:grid16lin_err_1}}
	\subfigure[]{\includegraphics[width=.42\columnwidth]{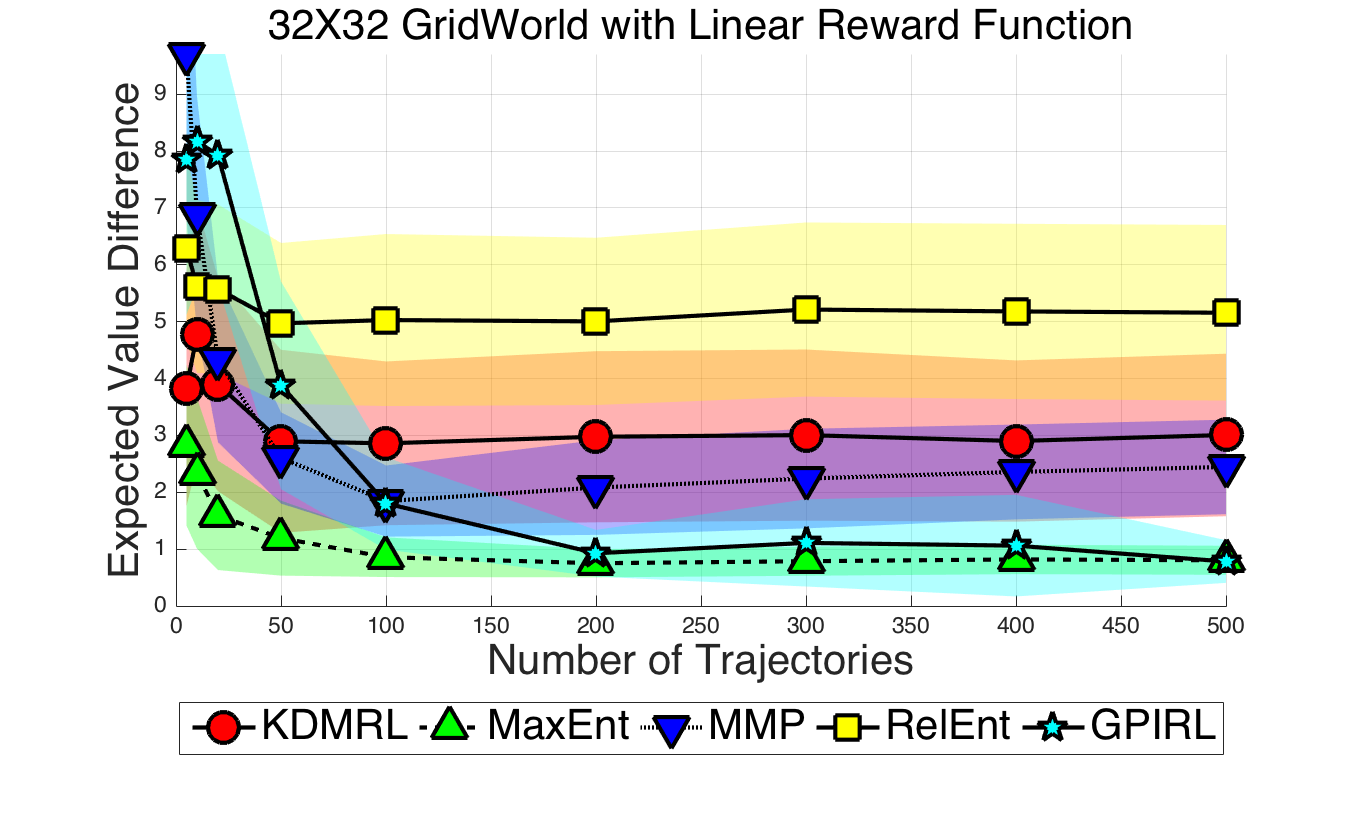} \label{fig:grid32lin_err_1}}
	\subfigure[]{\includegraphics[width=.42\columnwidth]{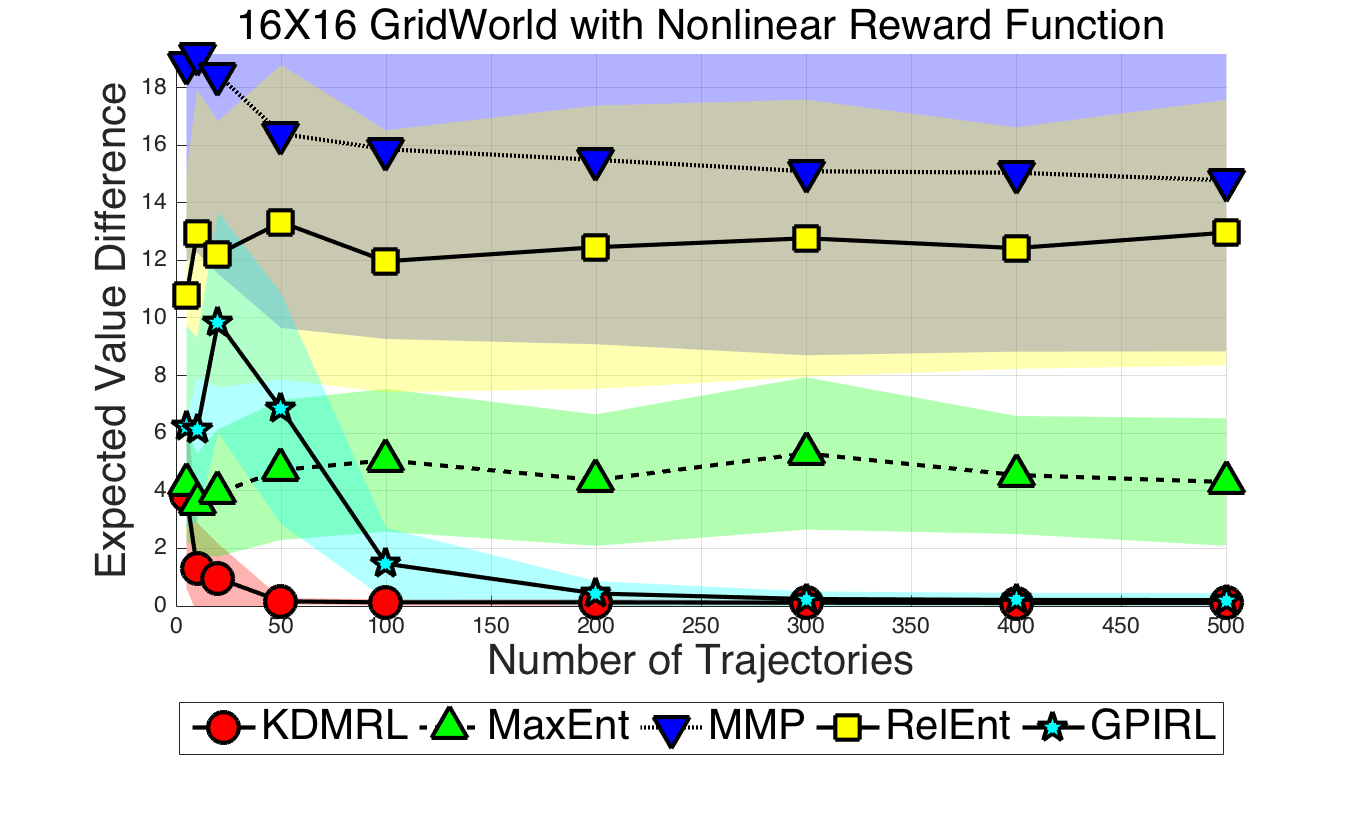} \label{fig:grid16nonlin_err_2}}
	\subfigure[]{\includegraphics[width=.42\columnwidth]{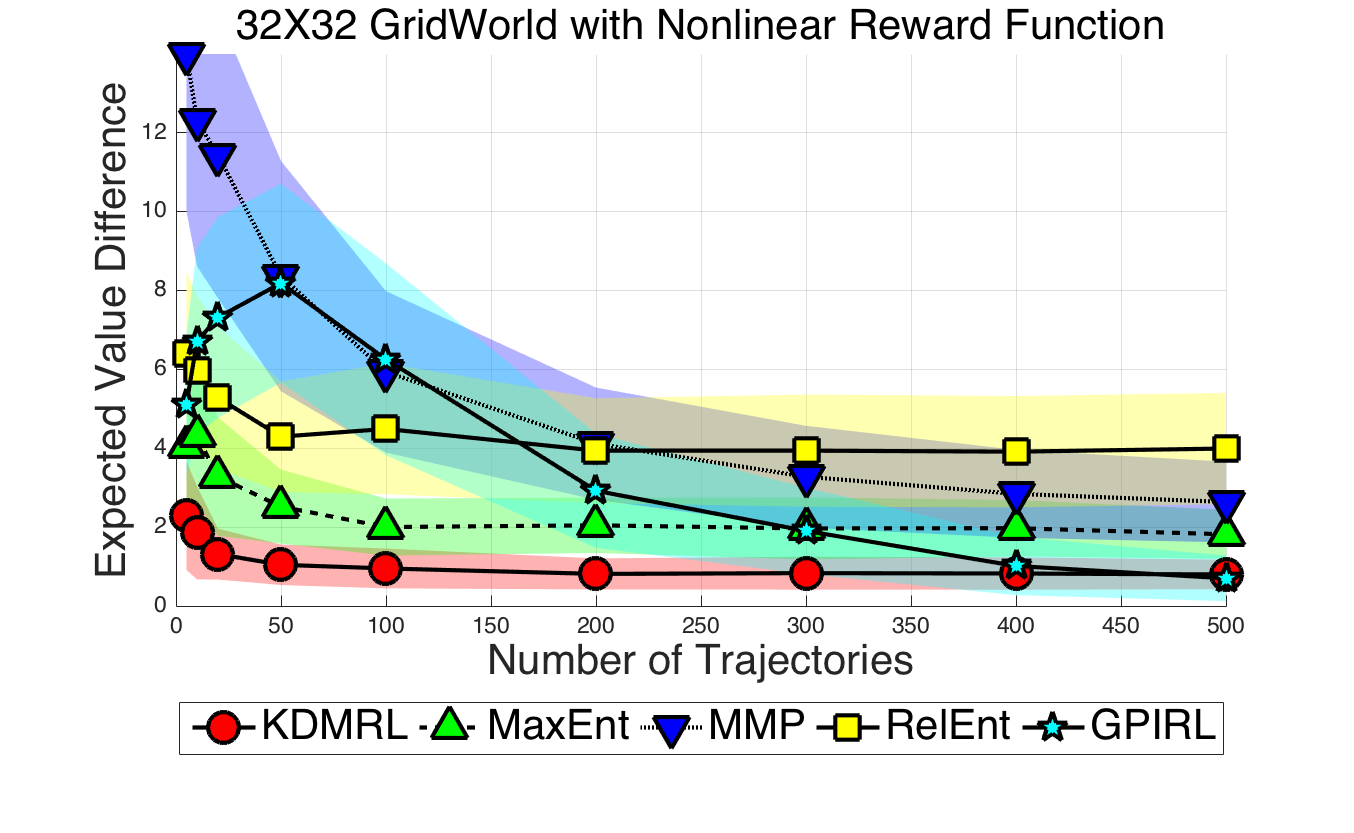} \label{fig:grid32nonlin_err_2}}
	\subfigure[]{\includegraphics[width=.42\columnwidth]{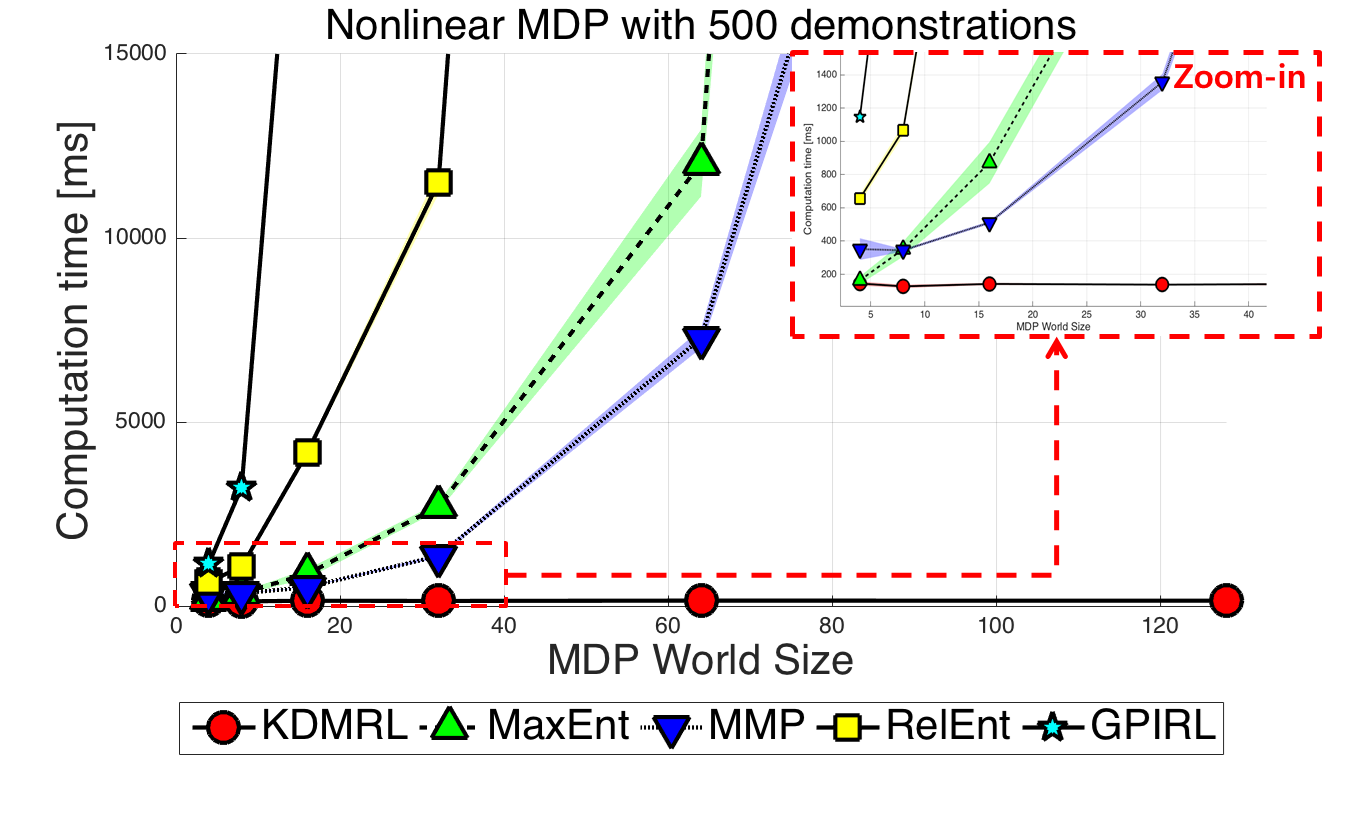} \label{fig:grid16lin_time_1}}
	\subfigure[]{\includegraphics[width=.42\columnwidth]{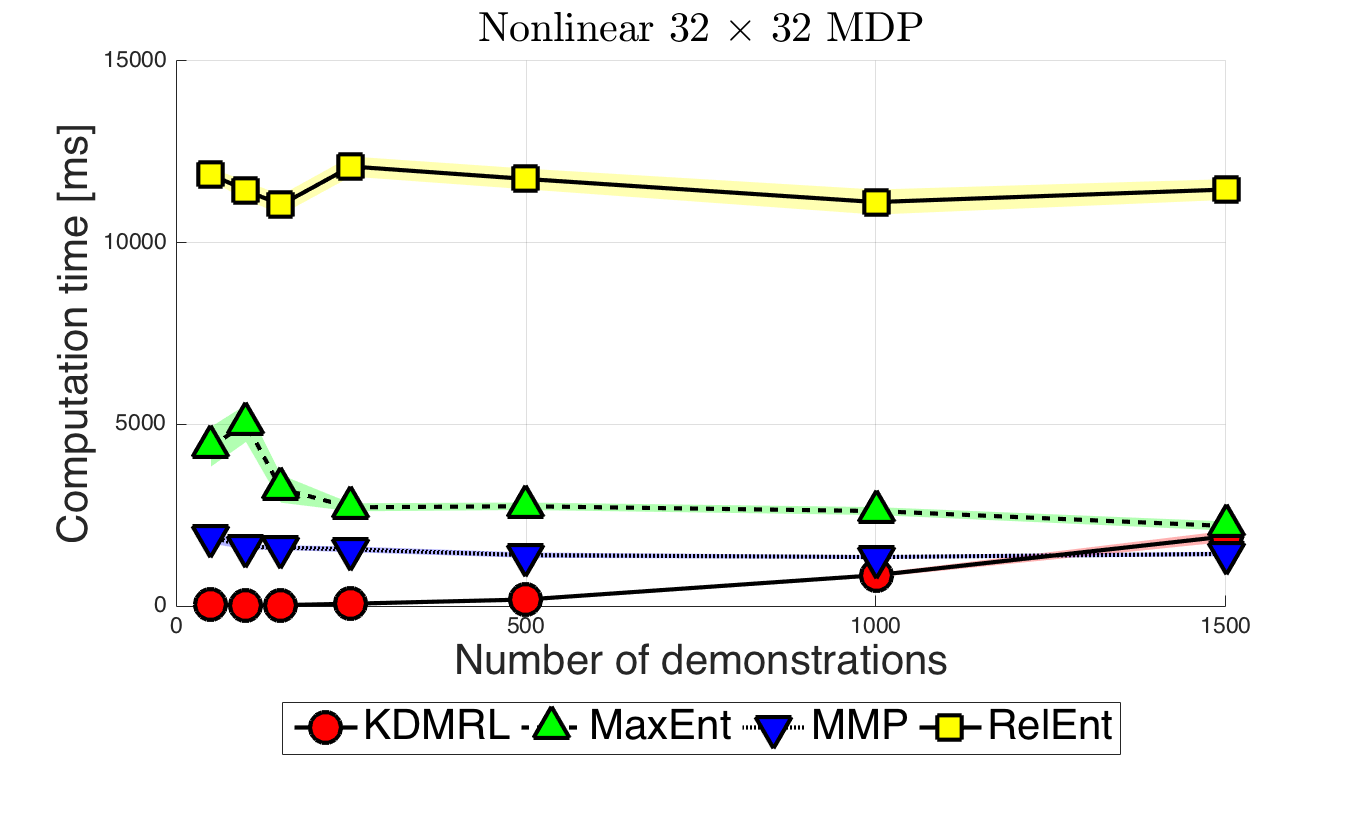} \label{fig:grid32lin_time_2}}
	\caption{
		Expected value difference of different IRL methods,
		where $50\%$ confidence intervals are shown in shaded colors. 
		(a) $16 \times 16$ grid world with a linear reward function. 
		(b) $32 \times 32$ grid world with a linear reward function.
		(c) $16 \times 16$ grid world with a nonlinear reward function. 
		(d) $32 \times 32$ grid world with a nonlinear reward function. 
		(e) Computation times at different MDP world sizes. 
		(f) Computation times at different numbers of demonstrations.
		}
	\label{fig:griderr}
\end{figure}

% Linear setting 
In the first set of experiments, $\{ \exp(-\norm{s-c_i}_2^2) \}_{i=1}^{k}$
is given as an observation or feature
making  the reward function as a linear function of observed features
(linear reward function). 
%In the linear reward setting, we make $10$ different reward maps
%and collected $5$ different set of trajectories per reward map
%to come up with $50$ different scenarios. 
% Nonlinear setting 
In the second set of experiments, features are 
defined as $\exp(-\norm{s-y_j}^2_2)$
where $y_i$ is $25$ equidistantly deployed points. 
In other words, the reward function is a nonlinear function 
of features (nonlinear reward function). 
We further separate the experiment into two grid worlds consist of 
$16 \times 16$ and $32 \times 32$ grid cells
resulting four different sets of experiments.
In both linear and nonlinear reward settings, 
we make $10$ different reward maps
and collected $5$ different sets of trajectories per reward map
to come up with $50$ different scenarios. 
% Settings
The trajectories of the experts are collected by solving the underlying MDP,
where lengths of trajectories are $8$ for $16 \times 16$ grid world
and $16$ for $32 \times 32$ grid world with discount factor of $0.95$. 
In all experiments, the number of peaks are $8$. 
For KDMRL, we set $\delta = 0.75$ for all experiments.

%% The KDMRL is compared with MaxEnt, MMP, RelEnt, and GPIRL 
%% %MaxEnt and MMP are model-base linear IRL method and GPIRL
%% %is a model-based nonlinear IRL method which often shows the 
%% %state of art performance on moderate number of demonstrations ($<5000$). 
%% %RelEnt is a model-free IRL method. 
%% where we vary the leverage of each training data by
%% modifying $\gamma$ in (\ref{eqn:kde2}) to be 
%% $\delta^{T-t}$ where $\delta = 0.75$ controls the leverage of each training
%% data, $T$ is the length of the trajectory, and $t$ is a time step. .
%% %This can be interpreted as a soft burn-in Monte Carlo Markov Chain (MCMC). 
%% %Once the policy function is fixed, an MDP becomes a Markov chain (MC)
%% %where the object of the DMRL is to estimate the stationary distribution 
%% %of the MC. However, due to the effect of the initial state distribution, 
%% %we increase the importance of latter (bigger time steps) demonstrations. 

% Results 
The expected value differences (EVDs) of IRL methods 
as a function of the number of trajectories are shown in 
Figure \ref{fig:grid16lin_err_1}--\ref{fig:grid32nonlin_err_2}. 
In the $16\times16$ linear reward setting, 
KDMRL outperforms
linear-based IRL methods, MaxEnt, MMP, and RelEnt
and shows performance comparable to GPIRL. 
However, in the $32 \times 32$ linear reward setting, 
model-based MaxEnt and MMP show better performance 
compared to KDMRL. 
But KDMRL still outperforms RelEnt, which is 
the only compared model-free IRL method. 
We believe that this is mainly due to two reasons. 
The first comes from the lack of world dynamics and 
the second comes from the limitation of a kernel-based method
as it is not appropriate for modeling a linear function (unless
a linear kernel function is used). 
% {\bf [Need to verify with linear DMRL.]}
% nonlinear 
On the other hand, in the nonlinear reward cases, 
KDMRL outperforms all linear IRL methods. 
KDMRL also outperforms GPIRL for both $16\times16$ and $32\times32$
cases when the number of samples is less than $200$
and shows comparable performance as more samples are used.

The time complexities are shown in
Figure~\ref{fig:grid16lin_time_1}--\ref{fig:grid32lin_time_2}. 
The computation time of KDMRL is less than $150$ms in $32\times32$
grid world on $2.2$GHz quad core processors. 
While the computation time increases linearly to the number of demonstrations, 
it is independent from the size of the state space as it does not solve an MDP
as its subroutine. 
This property plays a significant role in forthcoming experiments 
where the state space is much bigger. 

%%%%%%%%%%%%%%%%%%%%%%%%%%%%%%%%
% 									Track Driving Simulation
%%%%%%%%%%%%%%%%%%%%%%%%%%%%%%%%
\subsection{Track Driving Experiments}

%\begin{figure}[!t] \centering
%	\subfigure[]{\includegraphics[width=.42\columnwidth]
%		{figures/fig_bmrl_safe_driving_mode_xy_02} \label{fig:xy}}
%	\subfigure[]{\includegraphics[width=.45\columnwidth]
%		{figures/fig_bmrl_safe_driving_mode_cdist_avel} \label{fig:grid32lin_time}}
%	\subfigure[]{\includegraphics[width=.45\columnwidth]
%		{figures/fig_bmrl_safe_driving_mode_cdist_lanedeg} \label{fig:grid16lin_time}}
%	\subfigure[]{\includegraphics[width=.45\columnwidth]
%		{figures/fig_bmrl_safe_driving_mode_cdist_lanedev} \label{fig:grid32lin_time}}
%	\subfigure[]{\includegraphics[width=.45\columnwidth]
%		{figures/fig_bmrl_safe_driving_mode_ldist_avel} \label{fig:grid32lin_time}}
%	\subfigure[]{\includegraphics[width=.45\columnwidth]
%		{figures/fig_bmrl_safe_driving_mode_rdist_avel} \label{fig:grid32lin_time}}
%	\caption{
%		(a) Spatial densities of the trajectories of expert demonstrations (top) and resulting 
%		demonstrations from the trained reward (bottom). 
%		Feature densities of expert demonstrations (left) and resulting demonstrations from 
%		the trained reward (right) of
%		(b) center front distance  and angular velocity
%		(c) center front distance  and lane deviation degree
%		(d) center front distance  and lane deviation distance
%		(e) left front distance and angular velocity 
%		(f) right front distance and angular velocity 
%		}
%	\label{fig:lin_vs_gp_vicon}
%\end{figure}

% EXP2! 
In this experiments, we move on to more realistic and 
practical scenario, \textit{track driving} experiments.
In particular, we aim to learn three different driving styles 
(safe driving, speedy Gonzales, and tailgating modes)
from human demonstrations, which is similar to the
experiments in \cite{Levine_12}. 
All three driving styles have common objective of maintaining 
the center of the lane while avoiding collision with other cars. 
However, the safe driving and speedy Gonzales modes avoid cars in
front while the tailgating mode follows closely behind another car. 
In the safe driving and speedy Gonzales modes, a car
moves at $36km/h$ and $72km/h$, respectively.

The proposed method, KDMRL, is compared 
with MaxEnt, GPIRL, and RelEnt. 
MMP is not compared as it shows poor performance on 
nonlinear grid world experiments.
In the implementation of KDMRL, 
we define the inducing set as a union of training inputs with 
$1,000$ randomly sampled inputs similar to
\cite{levine2011nonlinear}, where the hyperparameters 
are selected using a simple median trick \cite{Dai_14}.\footnote{ 
While it is possible to fine-tune the hyperarameters
as (\ref{eqn:dmrl4}) is differentiable with respect to the 
hyperparameters, we omit this step as a simple median trick 
works well for this case.}

\begin{figure}[!t] \centering
	\subfigure[]{\includegraphics[width=.6\columnwidth]
		{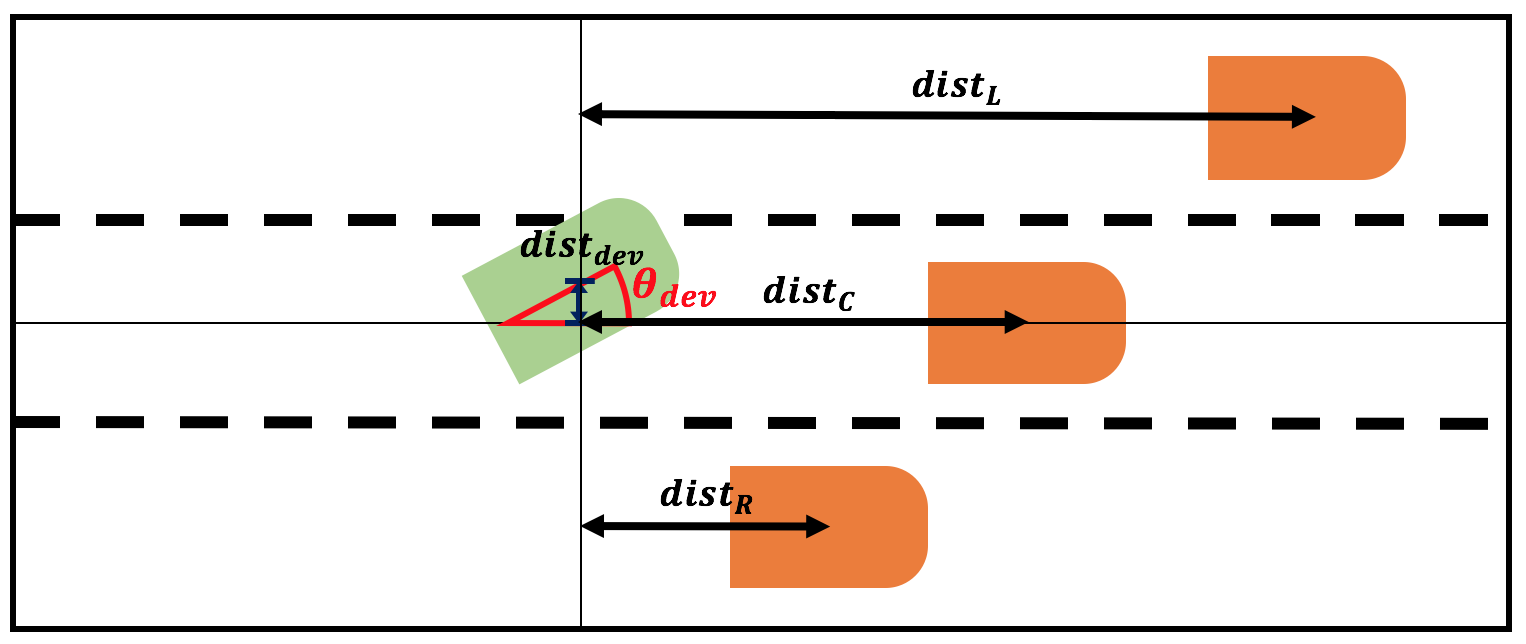} \label{fig:trackdesc}}
	\subfigure[]{\includegraphics[width=.25\columnwidth]
		{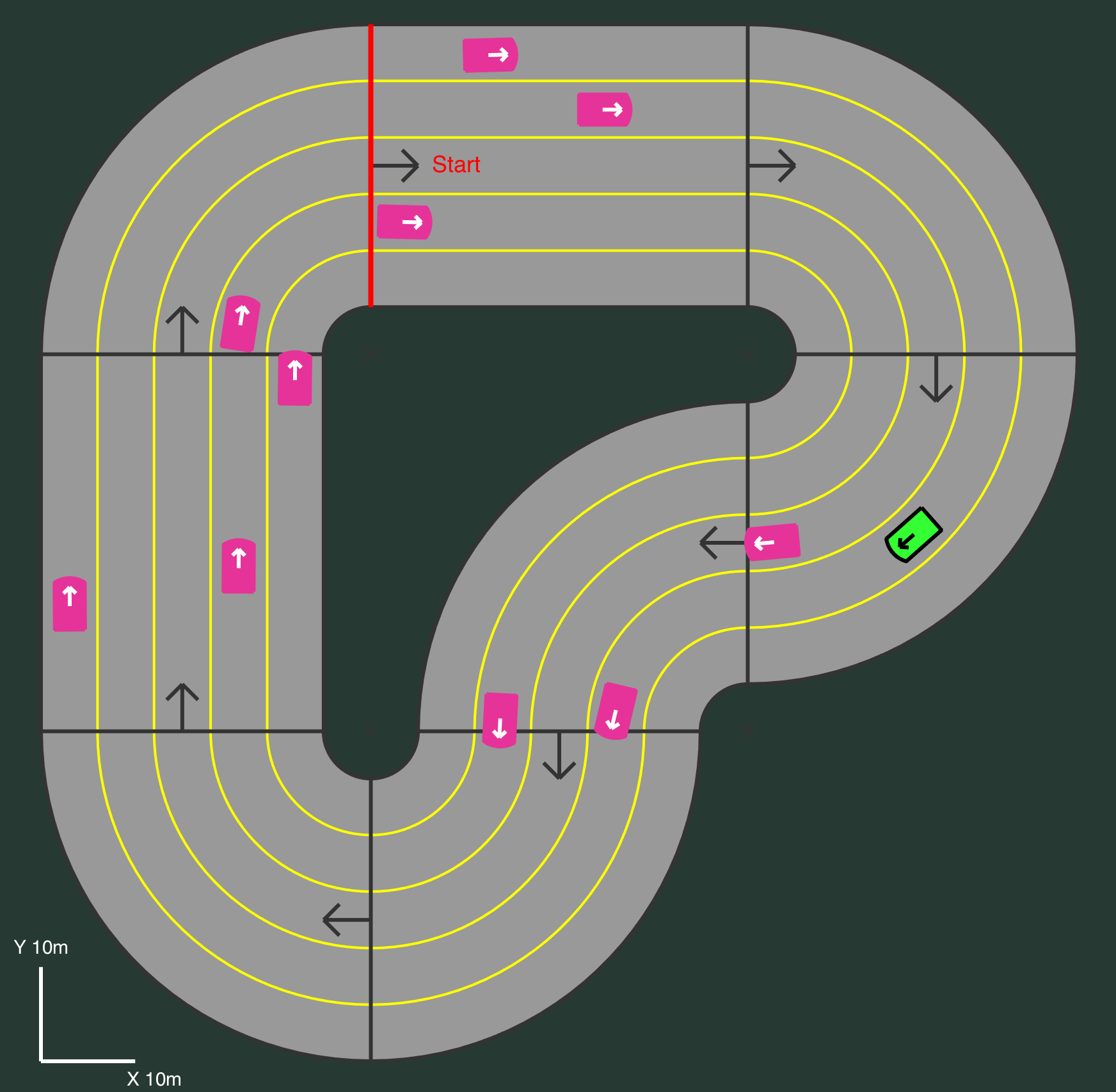} \label{fig:tracklarge}}
	\caption{
		(a) Descriptions of features used in the track driving experiments. 
		(b) Large track. 
		}
	\label{fig:lin_vs_gp_vicon}
\end{figure}

% The hardness of manually designing the reward function 
%The reward function is hard to design manually as it contains two
%conflicting terms; one trying the remain at the center of the road
%while the other try to avoid collisions. 
%In fact, we failed to manually design an appropriate reward function 
%that results desired driving behaviors
%which emphasizes the importance of IRL approaches. 

% State and feature representation 
The state and action consist of the pose
($x$, $y$, and $\theta$)
of a car following a unicycle dynamic model
% which is often used to model the dynamics of a simple car \cite{Lee01_tracking}
and directional and angular velocities ($v, \, w$), respectively. 
Both state and action spaces are discretizes for 
model based IRL methods by dividing the state space into 
$240$ states,
$3$ for vertical axis, $10$ for horizontal axis, 
and $8$ for heading, and the action space is divided into $9$ actions, 
$3$ for directional and $3$ for angular velocities.
We used six dimensional features to represent the 
reward function: 
(1) the lane deviation distance, $dist_{dev}$, 
(2) the lane deviation degree, $\theta_{dev}$, 
the closest frontal distances to other car in the
(3) left lane, $dist_L$, (4) center lane, $dist_C$, and 
(5) right lane, $dist_R$, 
and (6) the directional velocity, $v$, as depicted in  
Figure \ref{fig:trackdesc}.

% Demonstration gathering 
The demonstrations are manually collected by human demonstrators
with a simple straight road configuration with three lanes 
and randomly deployed cars as depicted in Figure \ref{fig:trackdesc}. 
For each driving mode, $10$ different settings of one to five randomly 
deployed car configurations are used. 
The demonstrations are collected by starting the car at each lane
resulting $30$ episodes for each mode. 
% The demonstrations are collected every $50$cm or $2$ degrees. 
% Exp1: on training scenario + How we control 
Once a reward function is trained, we use a
receding horizon controller to control the car.
%In particular, we sampled pre-defined $20$ trajectories 
%and weighted average the starting control signals to produce
%the actual control output for the KDMRL where the weights
%are defined as the sum rewards 
% collected while traversing the sampled trajectories. 
%This approach is closely related to the concept of 
%path integral control \cite{theodorou2010generalized}.
%Unlike the original path integral control which defines
%the weights with softmax function, we directly uses 
%the sum of rewards as the large scale variations often
%leads to numerical errors, e.g.,
%$\exp(1,000)$ is NaN in MATLAB.
%We also found that the resulting reward functions
%of other compared IRL methods to be unstable and 
%directly use the control that produces the maximum rewards
%to control the car. 

% Big table!!
\afterpage{\clearpage
\begin{landscape} 
\begin{table*}  \scriptsize \centering 
	\ra{1.2}
	\begin{tabular}{@{}crrrrcrrrrcrrrr@{}}
	{ \footnotesize (a) Results of trained scenarios} & & & & \\
	\toprule 
	& \multicolumn{4}{c}{Safe driving mode} & \phantom{abc}
	& \multicolumn{4}{c}{Speedy Gonzales} & \phantom{abc} 
	& \multicolumn{4}{c}{Tailgating mode} \\
	\cmidrule{2-5} \cmidrule{7-10} \cmidrule{12-15}
	& KDMRL & MaxEnt & GPIRL & RelEnt
	&& KDMRL & MaxEnt & GPIRL & RelEnt
	&& KDMRL & MaxEnt & GPIRL & RelEnt
	\\ \midrule
	Average collision ratio	
	& \bm{$0.0\%$} & $20\%$ & $23.3\%$ & $56.7\%$
	&& \bm{$0.0\%$} & $40\%$ & $20\%$ & $60\%$ 
	&& \bm{$0.0\%$} & $12.5\%$ & $78.3\%$ & $47.5\%$ \\
	$d_{var}(\tilde{P}_{X \times  Y}, \hat{P}_{X \times  Y})$
	& \bm{$0.074$} & $0.229$ & $0.18$  & $0.256$
	&& \bm{$0.085$} & $0.244$ & $0.163$ & $0.251$
	&& \bm{$0.152$} & $0.334$ & $0.313$ & $0.348$ \\
	$d_{var}(\tilde{P}_{dist_{C} \times w}, \hat{P}_{dist_{C} \times w})$ 
	& \bm{$0.106$} & $0.272$ & $0.303$  & $0.273$
	&& \bm{$0.134$} & $0.269$ & $0.284$ & $0.258$
	&& \bm{$0.232$} & $0.32$ & $0.569$ & $0.304$ \\ 
	$d_{var}(\tilde{P}_{dist_{C} \times dist_{dev}}, 
	\hat{P}_{dist_{C} \times dist_{dev}})$ 
	& \bm{$0.05$} & $0.665$ & $0.455$  & $0.687$
	&& \bm{$0.07$} & $0.648$ & $0.432$ & $0.73$
	&& \bm{$0.211$} & $0.705$ & $0.529$ & $0.741$ \\
	$d_{var}(\tilde{P}_{dist_{C} \times \theta_{dev}}, 
	\hat{P}_{dist_{C} \times \theta_{dev}})$
	& \bm{$0.046$} & $0.121$ & $0.149$  & $0.167$
	&& \bm{$0.055$} & $0.141$ & $0.14$ & $0.176$
	&& \bm{$0.197$} & $0.303$ & $0.179$ & $0.246$ \\
	$d_{var}(\tilde{P}_{dist_{R} \times w}, \hat{P}_{dist_{R} \times w})$
	& \bm{$0.104$} & $0.243$ & $0.328$  & $0.226$
	&& \bm{$0.119$} & $0.28$ & $0.289$ & $0.135$
	&& \bm{$0.15$} & $0.322$ & $0.543$ & $0.293$ \\
	$d_{var}(\tilde{P}_{dist_{L} \times w}, \hat{P}_{dist_{L} \times w})$ 
	& \bm{$0.113$} & $0.363$ & $0.252$  & $0.217$
	&& \bm{$0.128$} & $0.321$ & $0.261$ & $0.198$
	&& \bm{$0.151$} & $0.538$ & $0.555$ & $0.281$ \\
	\midrule
	Average variational distance
	& \bm{$0.082$} & $0.315$ & $0.278$  & $0.304$
	&& \bm{$0.098$} & $0.317$ & $0.261$ & $0.291$
	&& \bm{$0.182$} & $0.421$ & $0.448$ & $0.369$ \\
	\bottomrule
	\\
	{ \footnotesize (b) Results of transferred scenarios} & & & & 
	\\
	\toprule
	& \multicolumn{4}{c}{Safe driving mode} & \phantom{abc}
	& \multicolumn{4}{c}{Speedy Gonzales} & \phantom{abc} 
	& \multicolumn{4}{c}{Tailgating mode} \\
	\cmidrule{2-5} \cmidrule{7-10} \cmidrule{12-15}
	& KDMRL & MaxEnt & GPIRL & RelEnt
	&& KDMRL & MaxEnt & GPIRL & RelEnt
	&& KDMRL & MaxEnt & GPIRL & RelEnt
	\\ \midrule
	Average number of collision
	& \bm{$0.0$} & $2.2$ & $1.1$  & $1.6$
	&& \bm{$0.1$} & $2.2$ & $3.1$ & $4.0$
	&& \bm{$0.0$} & $2.2$ & $1.7$ & $2.2$ \\
	Absolute average lane deviation
	& \bm{$149.2$} & $1613.5$ & $1333.3$  & $1583.2$
	&& \bm{$590.3$} & $1613.5$ & $1196.6$ & $1503.0$
	&& \bm{$128.5$} & $1613.5$ & $1206.1$ & $1582.2$ \\
	Absolute average lane degree
	& \bm{$1.891$} & $8.462$ & $7.281$  & $7.941$
	&& \bm{$3.003$} & $8.462$ & $10.497$ & $13.431$
	&& \bm{$1.049$} & $8.462$ & $7.695$ & $10.374$ \\
	Average directional velocity
	& $9375.0$ & $15312.5$ & $11687.5$  & $9375.0$
	&& $17499.9$ & $15312.5$ & $17562.5$ & $17500.0$
	&& $9375.0$ & $15312.5$ & $9375.0$ & $9375.0$ \\
	Average number of lane change
	& \bm{$2.0$} & $23.6$ & $12.3$  & $12.8$
	&& \bm{$6.2$} & $23.6$ & $27.5$ & $34.9$
	&& \bm{$0.6$} & $23.6$ & $12.6$ & $17.7$ \\
%	$ d_{var}(\tilde{P}_{dist_{C} \times dist_{dev}}, 
%		\hat{P}_{dist_{C} \times dist_{dev}}) $ 
%	& \bm{$0.116$} & $0.698$ & $0.603$  & $0.726$
%	&& \bm{$0.389$} & $0.681$ & $0.532$ & $0.634$
%	&& $0.769$ & $0.717$ & \bm{$0.552$} & $0.702$ \\
%	$ d_{var}(\tilde{P}_{dist_{C} \times \theta_{dev}}, 
%		\hat{P}_{dist_{C} \times \theta_{dev}}) $
%	& \bm{$0.117$} & $0.184$ & $0.175$  & $0.199$
%	&& \bm{$0.153$} & $0.192$ & $0.254$ & $0.281$
%	&& $0.774$ & $0.261$ & \bm{$0.207$} & $0.259$ \\
	\bottomrule
	\end{tabular}
	\caption{(a) Average collision ratio and variational distances 
	between expert's distribution and empirical distribution
	obtained from running each algorithm on demonstrated
	scenarios. 
	(b) Experimental results on transferred scenarios. }
	\label{tbl:track_exp}
\end{table*}
\end{landscape}
}

% Quantifying 
We performed two sets of experiments to evaluate the performance of 
different IRL methods on track driving experiments. 
In the first set of experiments, the car is navigated 
in the road configuration where the demonstrations are collected. 
As we have the expert-controlled trajectories,
we compute the similarity measure between the reference
and resulting trajectories in terms of the variational distances. 
Followings are the six variational distances:
%{\bf [Describe metric below more clearly. Difficult tell what different subscripts mean.]}
%{\bf [It seems the measures described for the next experiment looks less confusing. You may define them here similarly.]}
%with the reasons for selecting such statistics:
\begin{enumerate} 
	\item $d_{var}(\tilde{P}_{X \times  Y}, \, \hat{P}_{X \times  Y})$
	to represent actual traversed distributions 
	\item $d_{var}(\tilde{P}_{dist_{C} \times w}, \, \hat{P}_{dist_{C} \times w})$ 
	to check collision avoiding behaviors
	\item $d_{var}(\tilde{P}_{dist_{C} \times dist_{dev}}, \,
	\hat{P}_{dist_{C} \times dist_{dev}})$ 
	to check center maintaining behaviors
	\item $d_{var}(\tilde{P}_{dist_{C} \times \theta_{dev}}, \,
	\hat{P}_{dist_{C} \times \theta_{dev}})$
	to check center maintaining behaviors
	\item $d_{var}(\tilde{P}_{dist_{R} \times w}, \, \hat{P}_{dist_{R} \times w})$ 
	to check approaching behaviors for tailgating mode
	\item $d_{var}(\tilde{P}_{dist_{L} \times w}, \, \hat{P}_{dist_{L} \times w})$ 
	to check approaching behaviors for tailgating mode
\end{enumerate}
where $d_{var}$ is a variational distance
and $\tilde{P}(\cdot)$ and $\hat{P}(\cdot)$
indicate the empirical distributions from
the reference trajectories and trajectories from the RHC, respectively. 
The average collision ratio is also computed by averaging 
the number of collided trajectories among $30$ trajectories for each mode. 

% Transfer
We further tested the controller on 
more complex and dynamic track environments with five lanes,
which is depicted in Figure \ref{fig:tracklarge}. 
This transfer can naturally be done as we used
a feature-based representation of the reward function. 
Each controller navigates the track for $60$ seconds 
where five to ten randomly deployed cars 
are moving at $30km/h$ maintaining its lane. 
Similar to the previous experiment, we computed the following
seven measures to evaluate the performances:
(1) Average number of collision, 
(2) Absolute average lane deviation distance,
(3) Absolute average lane deviation degree, 
(4) Average directional velocity, and
(5) Average number of lane changes.
%\begin{enumerate}
%	\item Average number of collision
%	\item Absolute average lane deviation distance
%	\item Absolute average lane deviation degree 
%	\item Average directional velocity
%	\item Average number of lane changes
%	\item $ d_{var}(\tilde{P}_{dist_{C} \times dist_{dev}}, 
%		\hat{P}_{dist_{C} \times dist_{dev}}) $ 
%	\item $ d_{var}(\tilde{P}_{dist_{C} \times \theta_{dev}}, 
%		\hat{P}_{dist_{C} \times \theta_{dev}}) $
%\end{enumerate}
%Note that the last two variational distances are computed by
%comparing the empirical distribution between two
%completely different track environments. 

% Results
The results of both trained and transferred scenarios are shown in 
Table \ref{tbl:track_exp}. 
(Also see the video submission for examples.)
First of all, in the trained scenarios, KDMRL outperforms all other methods
in terms of both collision ratio and variational distances with large margins. 
In particular, KDMRL navigates without any collision. 
GPIRL, which shows a superior performance on 
grid world experiments, shows second best performance on 
safe driving and speedy Gonzales mode. 
For the tailgating mode, however, 
it shows $78.3\%$ collision rate. 
It is mainly because it collides to a car in front 
while trying to drive behind other cars. 

\begin{figure}[t]
	\centering
	\includegraphics[width=0.55\columnwidth]{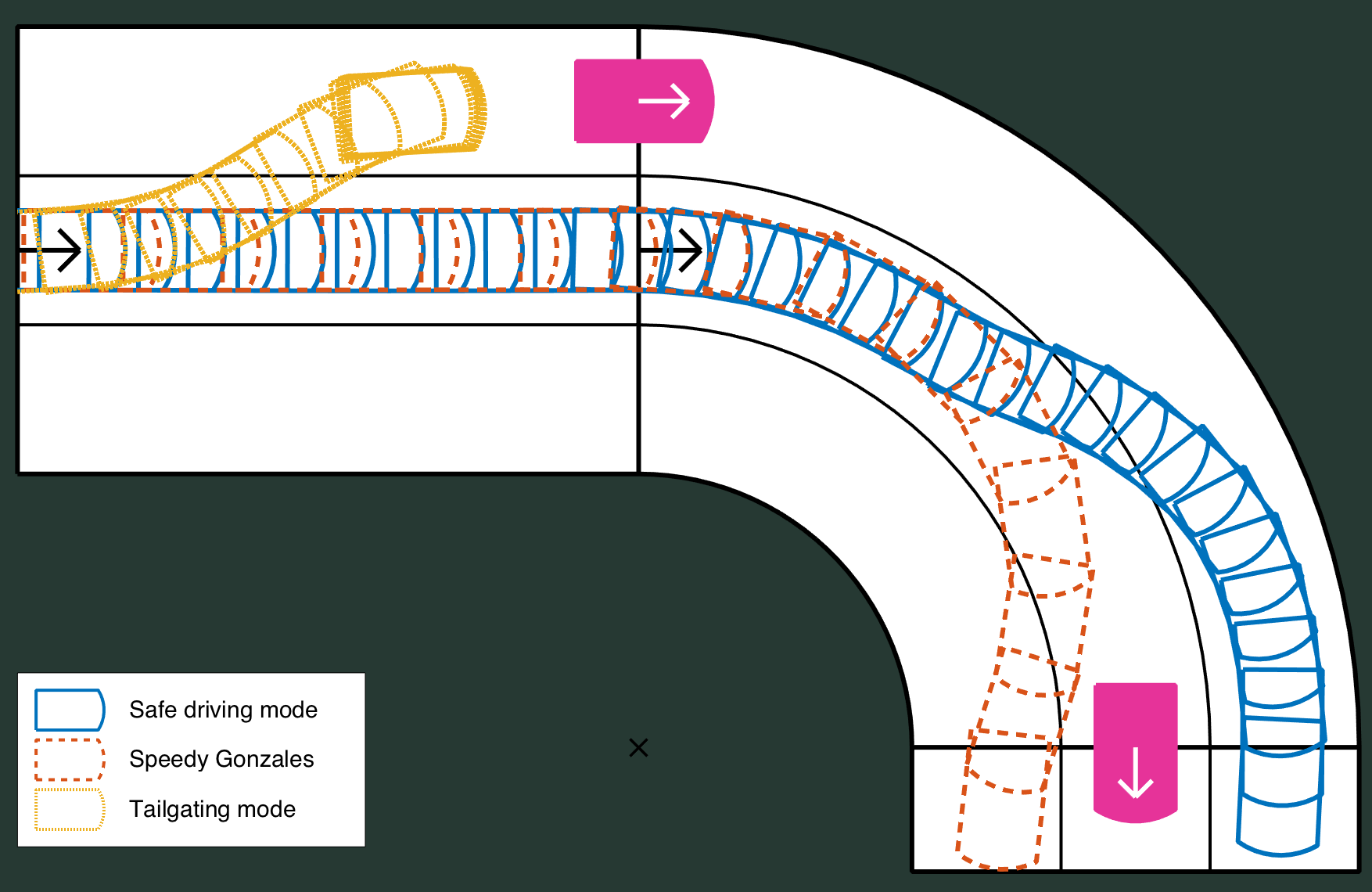}
	\caption{
		Snapshots of track driving experiments
		recorded every $0.2$ second with KDMRL on different 
		driving styles (green: safe driving mode,
		red: speedy Gonzales, and 
		yellow: tailgating mode).
	}
	\label{fig:trackres}
\end{figure}

In the transferred scenarios, 
the absolute average lane deviation and degree indicate 
how well a car maintain its center of the lane
and KDMRL outperforms other methods by a wide margin. 
However, this margin gets smaller from the safe driving mode
to speedy Gonzales as the car moves faster 
as shown by the average directional velocities. 
This stable driving behavior is also reflected by 
the number of lane changes, where KDMRL shows 
the minimum number. 
It is interesting to note that the average number of lane changes
in the tailgating mode is $0.6$. 
This is because, a car changes its lane only when a car in front
is in its left or right lane and will not change its lane once it
starts tailgating. 
%Of course, the number of lane changes gets larger from the safe driving mode
%to speedy Gonzales as the car moves faster. 
%However, KDMRL still shows the minimum number of lane changes. 
Resulting trajectories of three different driving styles 
with KDMRL are illustrated in Figure \ref{fig:trackres}.

%%%%%%%%%%%%%%%%%%%%%%%%%%%%%%%%
% 									Discussions
%%%%%%%%%%%%%%%%%%%%%%%%%%%%%%%%
\section{Discussions}

In this paper, we have considered the problem of solving a model-free 
nonlinear inverse reinforcement learning (IRL). 
Inspired by the fact that IRL can be viewed as a dual problem of
finding a reward function that matches the state-action distribution,
we first compute the empirical distribution of expert's trajectories
and find a reward function that matches the empirical distribution and
proposed density matching reward learning (DMRL).  
The sample complexity of the absolute difference between optimal and
estimated value of DMRL is also presented.  
Our method is further extended to a continuous state-action space
using a kernel-based function approximation method,
named kernel density matching reward learning (KDMRL).
The performance of KDMRL is extensively evaluated using two sets of
experiments: grid world and track driving experiments.
In the grid world experiments, KDMRL shows superior performance to
other IRL methods.
In the track driving experiments, KDMRL outperforms all other methods
in terms of variational distances. 
Furthermore, it was shown that the navigation with KDMRL is far more 
safe and stable while satisfying specific driving styles compared to
other methods.  

%%%%%%%%%%%%%%%%%%%%%%%%%%%%%%%%
% 									Assumptions
%%%%%%%%%%%%%%%%%%%%%%%%%%%%%%%%
\begin{subappendices} 
\renewcommand{\thesection}{\Alph{section}}
{\small  
\section{ } \label{appendix}
In this appendix, we introduce assumptions for 
Theorem \ref{thm:main2}.
Note that assumption \ref{ass:1} to \ref{ass:7}
are illustrated in \cite{Zeevi_97}, but we depict the assumptions here 
for completeness of the theorem. 
Assumption \ref{ass:8} is newly added. 
\begin{assumption}{}{}\label{ass:1}
	The random variables $\{ x_i\}_i^N$ are sampled 
	independent and identically distributed
	according to $f(x)$. 
\end{assumption}
\begin{assumption}{}{}\label{ass:2}
	The density estimator function $f_n^{\theta}$ is piecewise continuous 
	for each parameter $\theta$.
\end{assumption}
\begin{assumption}{}{}\label{ass:3}
	(a) $E[\log f(x)]$ exists and $|\log f_n^{\theta}(x)| \le m(x) ~ \forall \theta \in \Theta$
	for some $m(x)$
	where $m(x)$ is an integrable function with respect to $f$.
	(b) $E[\log(f/f_n^{\theta})]$ has a unique minimum. 
\end{assumption}
\begin{assumption}{}{}\label{ass:4}
	$\frac{\partial \log f_n^{\theta}(x)}{\partial \theta}$ 
	is integrable with respect to $x$
	and continuously differentiable. 
\end{assumption}
\begin{assumption}{}{}\label{ass:5}
	$|\frac{\partial^2 \log f_n^{\theta}(x)}{\partial \theta_i \partial \theta_j}|$
	and 
	$|\frac{\partial f_n^{\theta}(x)}{\partial \theta_i }\frac{\partial f_n^{\theta}(x)}{\partial \theta_j }|$
	is dominated by functions integrable with respect to $f$ $\forall x \in \mathcal{X}, ~ \theta \in \Theta$.
\end{assumption}
\begin{assumption}{}{}\label{ass:6}
	(a) $\theta^*$ is interior to $\Theta$. (b) $B(\theta^*)$ is nonsingular. 
	(c) $\theta^*$ is a regular point of $A(\theta)$. 
\end{assumption}
\begin{assumption}{}{}\label{ass:7}
	The convex model $f_n^{\theta} = \mathcal{G}_n$ obeys $\eta$ positivity requirement. 
\end{assumption}
\begin{assumption}{}{}\label{ass:8}
	The target density function $f \in \{ \sum_{i=1}^{\infty} \alpha_i \phi_{\rho}(\cdot ; \theta_i)  \}$
	where $\phi_{\rho}(\cdot ; \theta_i)$ is a basis density function.
\end{assumption}
}
\end{subappendices}

%%%%%%%%%%%%%%%%%%%%%%%%%%%%%%%%
% 									Bibliography
%%%%%%%%%%%%%%%%%%%%%%%%%%%%%%%%
\bibliographystyle{splncs03}
\bibliography{bib_bmrl}

\clearpage
\addtocmark[2]{Author Index} % additional numbered TOC entry
\renewcommand{\indexname}{Author Index}
\printindex
\clearpage
\end{document}